\documentclass{article}

\usepackage{times}
\usepackage{graphicx} % more modern
\usepackage{subfigure} 

\usepackage{natbib}

\usepackage{algorithm}
\usepackage{algorithmic}

\usepackage{hyperref}

\usepackage[accepted]{icml2016}
\usepackage{amsmath}
\usepackage{booktabs, multicol, multirow}
\usepackage{dirtree}

\icmltitlerunning{Deconstructing the Ladder Network Architecture}

\begin{document} 

\twocolumn[
\icmltitle{Deconstructing the Ladder Network Architecture}

\icmlauthor{Mohammad Pezeshki$^*$}{mohammad.pezeshki@umontreal.ca}
\icmlauthor{Linxi Fan$^\star$}{linxi.fan@columbia.edu}
\icmlauthor{Phil\'{e}mon Brakel$^*$}{pbpop3@gmail.com}
\icmlauthor{Aaron Courville$^{*\dagger}$}{aaron.courvile@umontreal.ca}
\icmlauthor{Yoshua Bengio$^{*\dagger}$}{yoshua.bengio@umontreal.ca}

\icmladdress{$^*$Universit\'{e} de Montr\'{e}al, $^\star$Columbia University, $^\dagger$CIFAR}

\icmlkeywords{boring formatting information, machine learning, ICML}

\vskip 0.3in
]

\begin{abstract} 
The Ladder Network is a recent new approach to semi-supervised learning that turned out to be very successful. 
While showing impressive performance, the Ladder Network has many components intertwined, whose contributions are not obvious in such a complex architecture. This paper presents an extensive experimental investigation of variants of the Ladder Network in which we replaced or removed individual components to learn about their relative importance.
For semi-supervised tasks, we conclude that 
the most important contribution is made by the lateral connections, followed by the application of noise, and the choice of what we refer to as the `combinator function'.
As the number of labeled training examples increases, the lateral connections and the reconstruction criterion
become less important, with most of the generalization improvement coming from the
injection of noise in each layer.
Finally, we introduce a combinator function that reduces test error rates on Permutation-Invariant MNIST to 0.57\% for the supervised setting, and to 0.97\% and 1.0\% for semi-supervised settings with 1000 and 100 labeled examples, respectively.
\end{abstract} 

\section{Introduction}
\label{Introduction}
Labeling data sets is typically a costly task and in many settings there are far more
unlabeled examples % (Let's not confuse everyone right off the bat here) or examples related to other tasks available 
than labeled ones.
Semi-supervised learning aims to improve the performance on some supervised
learning problem by using information obtained from both labeled and unlabeled examples.
Since the recent success of deep learning methods has mainly relied on supervised learning
based on very large labeled datasets, it is interesting to explore semi-supervised deep
learning approaches to extend the reach of deep learning to these settings. 

Since unsupervised methods for pre-training layers of neural networks were an
essential part of the first wave of deep learning methods
\citep{hinton2006fast,vincent2008extracting,bengio2009learning}, a natural next step
is to investigate how ideas inspired by Restricted Boltzmann Machine training
and regularized autoencoders can be used for semi-supervised learning.
Examples of approaches based on such ideas are the discriminative RBM \citep{larochelle2008}
and a deep architecture based on semi-supervised autoencoders that was used for
document classification \citep{ranzato2008semi}. 
More recent examples of approaches for semi-supervised deep learning are the
semi-supervised Variational Autoencoder \citep{kingma2014semi} and the Ladder
Network \citep{Rasmus2015} which obtained very impressive, state of the art results (1.13\% error)
on the MNIST
handwritten digits classification benchmark using just 100 labeled training examples.

The Ladder Network adds an unsupervised component to the supervised learning
objective of a deep feedforward network by treating this network as part of a
deep stack of denoising autoencoders or DAEs~\citep{Vincent-JMLR-2010-small} 
that learns to reconstruct each layer (including the input) based on a corrupted
version of it, using feedback from upper levels. The term 'ladder' refers to how
this architecture extends the stacked DAE in the way the feedback paths are formed.

This paper is focusing on the design choices that lead to the
Ladder Network's superior performance and tries to disentangle them empirically.
We identify some general properties of the model
that make it different from standard feedforward networks and compare various
architectures to identify those properties and design choices that are the most
essential for obtaining good performance with the Ladder Network. While the original authors of the Ladder Network paper explored some variants of their model already, we provide a thorough comparison of a large number of architectures controlling for both hyperparameter settings and data set selection.
Finally, we also introduce a variant of the Ladder Network that yields new state-of-the-art results for the Permutation-Invariant MNIST classification task in both semi- and fully- supervised settings.

\section{The Ladder Network Architecture}

In this section, we describe the Ladder Network Architecture\footnote{Please refer to \citep{Rasmus2015,valpola2014neural} for more detailed explanation of the Ladder Network architecture.}. Consider a dataset with $N$ labeled examples ${(x(1), y^*(1)), (x(2), y^*(2)), ..., (x(N), y^*(N))}$ and $M$ unlabeled examples ${x(N+1), x(N+2), ..., x(N+M)}$ where $M \gg N$. The objective is to learn a function that models $P(y|x)$ by using both the labeled examples and the large quantity of unlabeled examples. In the case of the Ladder Network, this function is a deep Denoising Auto Encoder (DAE) in which noise is injected into all hidden layers and the objective function is a weighted sum of the supervised Cross Entropy cost on the top of the encoder and the unsupervised denoising Square Error costs at each layer of the decoder. Since all layers are corrupted by noise, another encoder path with shared parameters is responsible for providing the clean reconstruction targets, i.e. the noiseless hidden activations (See Figure 1).

Through lateral skip connections, each layer of the noisy encoder is connected to its corresponding layer in the decoder. This enables the higher layer features to focus on more abstract and task-specific features. Hence, at each layer of the decoder, two signals, one from the layer above and the other from the corresponding layer in the encoder are combined.

Formally, the Ladder Network is defined as follows:
\begin{align}
\tilde{x}, \tilde{z}^{(1)}, ..., \tilde{z}^{(L)}, \tilde{y} &= \mbox{Encoder}_{noisy}(x),\\
x, z^{(1)}, ..., z^{(L)}, y &= \mbox{Encoder}_{clean}(x),\\
\hat{x}, \hat{z}^{(1)}, ..., \hat{z}^{(L)} &= \mbox{Decoder}(\tilde{z}^{(1)}, ..., \tilde{z}^{(L)}),
\end{align}
where Encoder and Decoder can be replaced by any multi-layer architecture such as a multi-layer perceptron in this case. The variables $x$, $y$, and $\tilde{y}$ are the input, the noiseless output, and the noisy output respectively. The true target is denoted as $y^*$. The variables $z^{(l)}$, $\tilde{z}^{(l)}$, and $\hat{z}^{(l)}$ are the hidden representation, its noisy version, and its reconstructed version at layer $l$. The objective function is a weighted sum of supervised (Cross Entropy) and unsupervised costs (Reconstruction costs).
\begin{align}\label{cost_func}
Cost &= -\Sigma_{n=1}^N\log P\big(\tilde{y}(n)=y^*(n) | x(n)\big) +\\ &\Sigma_{n=N+1}^{M} \Sigma_{l=1}^{L} \lambda_l\, \mbox{ReconsCost}(z^{(l)}(n),\hat{z}^{(l)}(n)).
\end{align}

Note that while the noisy output $\tilde{y}$ is used in the Cross Entropy term, the classification task is performed by the noiseless output $y$ at test time.

In the forward path, individual layers of the encoder are formalized as a linear transformation followed by Batch Normalization \citep{ioffe2015batch} and then application of a nonlinear activation function:
\begin{align}
\tilde{z}^{(l)}_{pre} &= W^{(l)} \cdot \tilde{h}^{(l-1)}, \label{eq:encoder_start} \\ 
\mu^{(l)} &= \mbox{mean}(\tilde{z}^{(l)}_{pre}),\\ 
\sigma^{(l)} &= \mbox{stdv}(\tilde{z}^{(l)}_{pre}),\\
\tilde{z}^{(l)} &= \frac{\tilde{z}^{(l)}_{pre} - \mu^{(l)}}{\sigma^{(l)}} + \mathcal{N}(0, \sigma^2),\\
\tilde{h}^{(l)} &= \phi\big(\gamma^{(l)} (\tilde{z}^{(l)} + \beta^{(l)})\big) \label{eq:encoder_end}
\end{align}

where $\tilde{h}^{(l-1)}$ is the post-activation at layer $l-1$ and $W^{(l)}$ is the weight matrix from layer $l-1$ to layer $l$. Batch Normalization is applied to the pre-normalization $\tilde{z}_{\textit{pre}}^{(l)}$ using the mini-batch mean $\mu^{(l)}$ and standard deviation $\sigma^{(l)}$. The next step is to add Gaussian noise with mean $0$ and variance $\sigma^2$ to compute pre-activation $\tilde{z}^{(l)}$. The parameters $\beta^{(l)}$ and $\gamma^{(l)}$ are responsible for shifting and scaling before applying the nonlinearity $\phi(\cdot)$.
Note that the above equations describe the \emph{noisy} encoder.
If we remove noise ($\mathcal{N}(0, \sigma^2)$) and replace $\tilde{h}$ and $\tilde{z}$ with $h$ and $z$ respectively, we will obtain the noiseless version of the encoder.

At each layer of the decoder in the backward path, the signal from the layer $\hat{z}^{(l+1)}$ and the noisy signal $\tilde{z}^{(l)}$ are combined into the reconstruction $\hat{z}^{(l)}$ by the following equations:
\begin{align}
u^{(l+1)}_{pre} &= V^{(l)} \cdot \hat{z}^{(l+1)}, \label{eq:decoder_start} \\ 
\mu^{(l+1)} &= \mbox{mean}(u^{(l+1)}_{pre}),\\
\sigma^{(l+1)} &= \mbox{stdv}(u^{(l+1)}_{pre}),\\
u^{(l+1)} &= \frac{u^{(l+1)}_{pre} - \mu^{(l+1)}}{\sigma^{(l+1)}},\\
\hat{z}^{(l)} &= g(\tilde{z}^{(l)}, u^{(l+1)}) \label{eq:decoder_end}
\end{align}

where $V^{(l)}$ is a weight matrix from layer $l+1$ to layer $l$. We call the function $g(\cdot,\cdot)$ the \textit{combinator function} as it combines the vertical $u^{(l+1)}$ and the lateral $\tilde{z}^{(l)}$ connections in an element-wise fashion. The original Ladder Network proposes the following design for $g(\cdot,\cdot)$, which we call the \textit{vanilla combinator}:
	\begin{align} \label{vanilla_comb}
	\begin{split}
	g(\tilde{z}^{(l)}, u^{(l+1)}) &= b_0 + w_{0z} \odot \tilde{z}^{(l)} +\\ 
    &w_{0u} \odot u^{(l+1)} + w_{0zu} \odot \tilde{z}^{(l)} \odot u^{(l+1)} +\\
    &w_{\sigma}\odot\mbox{Sigmoid}(b_1 + w_{1z} \odot \tilde{z}^{(l)} +\\ 
    &w_{1u} \odot u^{(l+1)} + w_{1zu} \odot \tilde{z}^{(l)} \odot u^{(l+1)}),
	\end{split}
	\end{align}
    
where $\odot$ is an element-wise multiplication operator and each per-element weight is initialized as:
\begin{align} \label{vanilla_init}
	\begin{cases}
    	w_{\{0, 1\}z} &\leftarrow 1 \\
        w_{\{0, 1\}u} &\leftarrow 0 \\
        w_{\{0, 1\}zu}, b_{\{0, 1\}} &\leftarrow 0 \\
        w_{\sigma} &\leftarrow 1
	\end{cases}
\end{align}
In later sections, we will explore alternative initialization schemes on the vanilla combinator.
Finally, the $\mbox{ReconsCost}(z^{(l)},\hat{z}^{(l)})$ in equation (\ref{cost_func}) is defined as the following:
\begin{align} \label{RC}
	\mbox{ReconsCost}(z^{(l)},\hat{z}^{(l)}) = ||\frac{\hat{z}^{(l)} - \mu^{(l)}}{\sigma^{(l)}}-z^{(l)}||^2.
\end{align}

where $\hat{z}^{(l)}$ is normalized using $\mu^{(l)}$ and $\sigma^{(l)}$ which are the \emph{encoder}'s sample mean and standard deviation statistics of the current mini batch, respectively. The reason for this second normalization is to cancel the effect of unwanted noise introduced by the limited batch size of Batch Normalization.

\begin{figure*}[htb]
  \centering
    \includegraphics[width=0.78\textwidth]{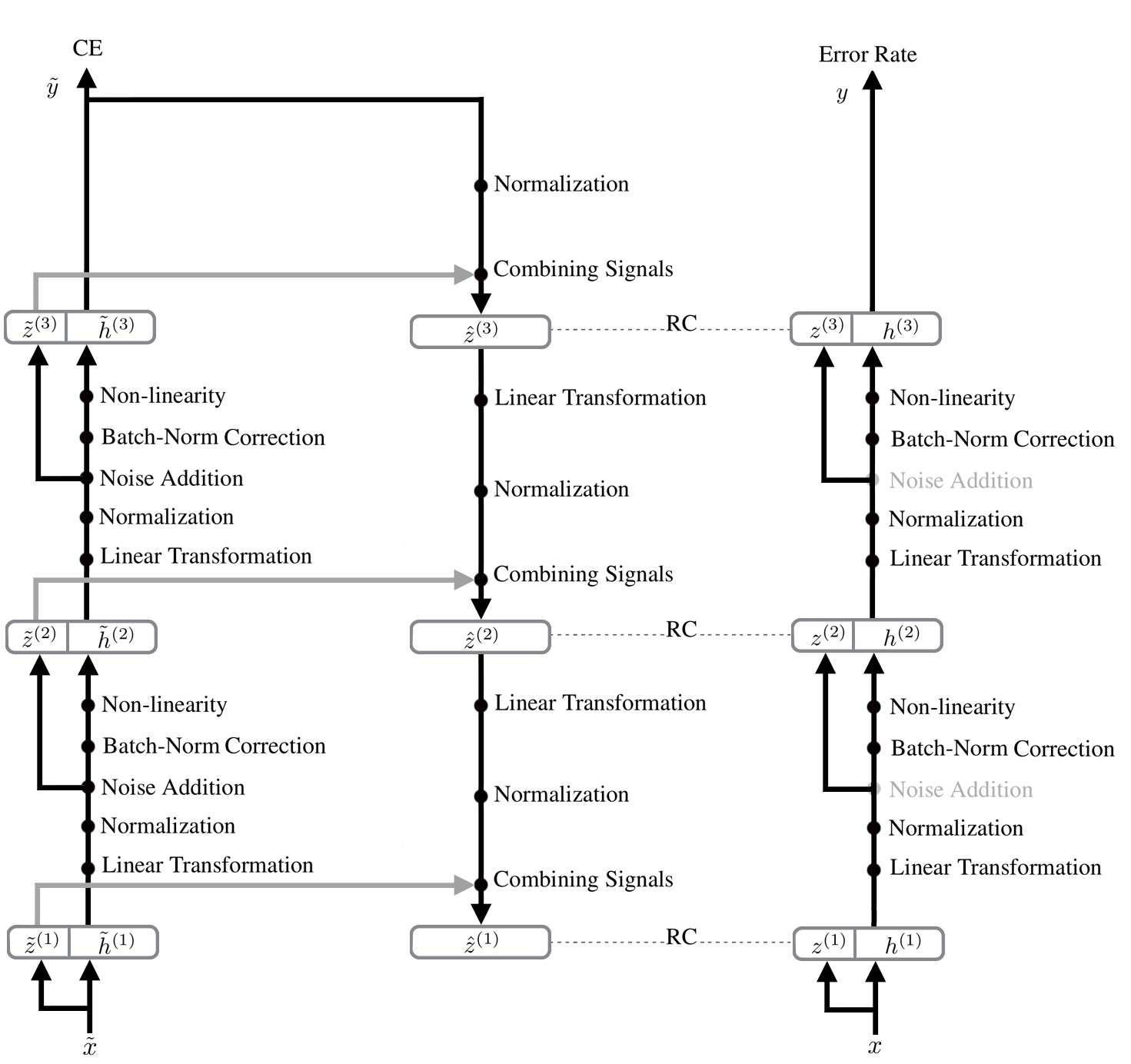}
    \caption{The Ladder Network consists of two encoders (on each side of the figure) and one decoder (in the middle). At each layer of both encoders (equations \ref{eq:encoder_start} to \ref{eq:encoder_end}), $z^{(l)}$ and $\tilde{z}^{(l)}$ are computed by applying a linear transformation and normalization on $h^{(l-1)}$ and $\tilde{h}^{(l-1)}$, respectively. The noisy version of the encoder (left) has an extra Gaussian noise injection term. Batch normalization correction ($\gamma^l, \beta^l$) and non-linearity are then applied to obtain $h^{(l)}$ and $\tilde{h}^{(l)}$.
At each layer of the decoder, two streams of information, the lateral connection $\tilde{z}^{(l)}$ (gray lines) and the vertical connection $u^{(l+1)}$, are required to reconstruct $\hat{z}^{(l)}$ (equations \ref{eq:decoder_start} to \ref{eq:decoder_end}).
Acronyms CE and RC stand for Cross Entropy and Reconstruction Cost respectively. The final objective function is a weighted sum of all Reconstruction costs and the Cross Entropy cost.}
\end{figure*}

\section{Components of the Ladder Network}

Now that the precise architecture of the Ladder Network has been described in
details, we can identify a couple of important additions to the standard
feed-forward neural network architecture that may have a pronounced impact on the
performance. A distinction can also be made between those design choices that
follow naturally from the motivation of the ladder network as a deep
autoencoder and those that are more ad-hoc and task specific.

The most obvious addition is the extra reconstruction cost for every hidden layer and the input layer. While it is clear that the reconstruction cost provides an
unsupervised objective to harness the unlabeled examples, it is not clear how important the penalization is for each layer and what role it plays for fully-supervised tasks.

A second important change is the addition of Gaussian noise to the input and
the hidden representations. While adding noise to the first layer is a part of denoising autoencoder training, it is again not clear whether it is necessary to add this noise at every layer or not.
We would also like to know if the noise helps by making the reconstruction task
nontrivial and useful or just by regularizing the feed-forward network in a similar
way that noise-based regularizers like dropout \citep{srivastava2014} and
adaptive weight noise \citep{graves2011practical} do.

Finally, the lateral skip connections are the most notable deviation from the standard denoising autoencoder architecture. The way the vanilla Ladder Network combines the lateral stream of information $\tilde{z}^{(l)}$ and the downward stream of information $u^{(l+1)}$ is somewhat unorthodox. For this reason, we have conducted extensive experiments on both the importance of the lateral connections and the precise choice for the function that combines the lateral and downward information streams (which we refer to as the \emph{combinator function}).

\section{Experimental Setup}
In this section, we introduce different variants of the Ladder Architecture and describe our experiment methodology. Some variants are derived by removal of one component of the model while other variants are derived by the replacement of that component with a new one. This enables us to isolate each component and observe its effects while other components remain unchanged. Table \ref{itm:variants} depicts the hierarchy of the different variants and the baseline models.

\begin{table}[h!]
\caption{Schematic ordering of the different models. All the variants of the \textsc{Vanilla} model are derived by either \textit{removal} or \textit{replacement} of a single component. The \textsc{Baseline} is a multi-layer feedforward neural networks with the same number of layers and units as the vanilla model.}

\label{itm:variants}
\dirtree{%
.1 Models.
.2 \textsc{Vanilla}.
.3 Removal of a component.
.4 Noise variants.
.5 \textsc{FirstNoise}.
.5 \textsc{FirstRecons}.
.5 \textsc{FirstN\&R}.
.5 \textsc{NoLateral}.
.4 Vanilla Combinator variants.
.5 \textsc{NoSig}.
.5 \textsc{NoMul}.
.5 \textsc{Linear}.
.3 Replacement of a component.
.4 Vanilla Combinator Initialization.
.5 \textsc{RandInit}.
.5 \textsc{RevInit}.
.4 MLP Combinators.
.5 \textsc{MLP}.
.5 \textsc{Augmented-MLP}.
.4 Gaussian Combinators.
.5 \textsc{Gaussian}.
.5 \textsc{GatedGauss}.
.2 Baseline models.
.3 Feedforward network.
.3 Feedforward network + noise.
}
\end{table}

\subsection{Variants derived by removal of a component}

\subsubsection{Noise Variants}

Different configurations of noise injection, penalizing reconstruction errors, and the lateral connection removal suggest four different variants: 

\begin{itemize}
\item Add noise only to the first layer (\textsc{FirstNoise}).

\item Only penalize the reconstruction at the first layer (\textsc{FirstRecons}), i.e. $\lambda^{(l \ge 1)}$ are set to $0$.

\item Apply both of the above changes: add noise and penalize the reconstruction only at the first layer (\textsc{FirstN\&R}).

\item Remove all lateral connections from \textsc{FirstN\&R}. Therefore, equivalent to to a denoising autoencoder with an additional supervised cost at the top, the encoder and the decoder are connected only through the topmost connection.
We call this variant \textsc{NoLateral}.
\end{itemize}

\subsubsection{Vanilla combinator variants} \label{sec:vanilla_variants}

We try different variants of the vanilla combinator function that combines the two streams of information from the lateral and the vertical connections in an unusual way. As defined in equation \ref{vanilla_comb}, the output of the vanilla combinator depends on $u$, $\tilde{z}$, and $u \odot \tilde{z}$ \footnote{For simplicity, subscript $i$ and superscript $l$ are implicit from now on.}, which are connected to the output via two paths, one linear and the other through a sigmoid non-linearity unit.

The simplest way of combining lateral connections with vertical connections is to simply add them in an element-wise fashion, similar to the nature of skip-connections in a recently published work on Residual Learning \citep{he2015deep}. We call this variant \textsc{Linear} combinator function. We also derive two more variants \textsc{NoSig} and \textsc{NoMult} in the way of stripping down the vanilla combinator function to the simple \textsc{Linear} one:

\begin{itemize}
\item Remove sigmoid non-linearity (\textsc{NoSig}). The corresponding per-element weights are initialized in the same way as the vanilla combinator.

\item Remove the multiplicative term $\tilde{z} \odot u$ (\textsc{NoMult}).

\item Simple linear combination (\textsc{Linear})
\begin{align}
g(\tilde{z}, u) = b + w_u \odot u + w_z \odot \tilde{z}
\end{align}
where the initialization scheme resembles the vanilla one in which $w_{z}$ is initialized to one while $w_{u}$ and $b$ are initialized to zero.
\end{itemize}

\subsection{Variants derived by replacement of a component}
\subsubsection{Vanilla combinator initialization}
Note that the vanilla combinator is initialized in a very specific way (equation \ref{vanilla_init}), which sets the initial weights for lateral connection $\tilde{z}$ to 1 and vertical connection $u$ to 0. This particular scheme encourages the Ladder decoder path to learn more from the lateral information stream $\tilde{z}$ than the vertical $u$ at the beginning of training. 

We explore two variants of the initialization scheme:
\begin{itemize}
	\item Random initialization (\textsc{RandInit}): all per-element parameters are randomly initialized to $\mathcal{N}(0, 0.2)$.
    \item Reverse initialization (\textsc{RevInit}): all per-element parameters $w_{\{0, 1\}z}$, $w_{\{0, 1\}zu}$, and $b_{\{0, 1\}}$ are initialized to zero while $w_{\{0, 1\}u}$ and $w_{\sigma}$ are initialized to one.
\end{itemize}
\subsubsection{Gaussian combinator variants}
Another choice for the combinator function with a probabilistic interpretation is the \textsc{Gaussian} combinator proposed in the original paper about the Ladder Architecture \citep{Rasmus2015}. Based on the theory in the Section 4.1 of \citep{valpola2014neural}, assuming that both additive noise and the conditional distribution $P(z^{(l)}|u^{(l+1)})$ are Gaussian distributions, the denoising function is linear with respect to $\tilde{z}^{(l)}$. Hence, the denoising function could be a weighted sum over $\tilde{z}^{(l)}$ and a prior on $z^{(l)}$. The weights and the prior are modeled as a function of the vertical signal:
\begin{align}
   g(\tilde{z}, u) &= \nu(u) \odot \tilde{z} + (1-\nu(u)) \odot \mu(u), \label{gauss_comb}
\end{align}
in which
\begin{align}
	\mu(u)&=w_1\odot\text{Sigmoid}(w_2 \odot u + w_3) + w_4 \odot u + w_5, \label{gauss_mu}\\ 
    \nu(u)&=w_6\odot\text{Sigmoid}(w_7 \odot u + w_8) + w_9 \odot u + w_{10}. \label{gauss_v}
\end{align}
Strictly speaking, $\nu(u)$ is not a proper weight, because it is not guaranteed to be positive all the time. To make the Gaussian interpretation rigorous, we explore a variant that we call \textsc{GatedGauss}, where equations \ref{gauss_comb} and \ref{gauss_mu} stay the same but \ref{gauss_v} is replaced by:
\begin{align}
    \nu(u)&= \text{Sigmoid}(w_6 \odot u + w_7).   
\end{align}
\textsc{GatedGauss} guarantees that $0 < \nu(u) < 1$.
We expect that $\nu(u)_i$ will be close to 1 if the information from the lateral connection for unit $i$ is more helpful to reconstruction, and close to 0 if the vertical connection becomes more useful. The \textsc{GatedGauss} combinator is similar in nature to the gating mechanisms in other models such as Gated Recurrent Unit \citep{cho2014properties} and highway networks \citep{srivastava2015training}.

\subsubsection{MLP (Multilayer Perceptron) combinator variants}

We also propose another type of element-wise combinator functions based on fully-connected MLPs. 
We have explored two classes in this family. The first one, denoted simply as \textsc{MLP}, maps two scalars $[u, \tilde{z}]$ to a single output $g(\tilde{z}, u)$. We empirically determine the choice of activation function for the hidden layers. Preliminary experiments show that the Leaky Rectifier Linear Unit (LReLU) \citep{maas2013rectifier} performs better than either the conventional ReLU or the sigmoid unit. Our LReLU function is formulated as 
	\begin{align} \label{LReLU}
	\text{LReLU}(x) &= \begin{cases}
		x,\,\, \mbox{   if } x \ge 0,\\
		0.1\,x,\, \mbox{   otherwise}
		\end{cases}.
	\end{align}
We experiment with different numbers of layers and hidden units per layer in the \textsc{MLP}. We  present results for three specific configurations: $[4]$ for a single hidden layer of 4 units, $[2, 2]$ for 2 hidden layers each with 2 units, and $[2, 2, 2]$ for 3 hidden layers. 
For example, in the $[2, 2, 2]$ configuration, the \textsc{MLP} combinator function is defined as:
   \begin{align}
   \begin{split}
   g(\tilde{z}, u) &= W_3 \cdot \text{LReLU}\Big(\\&W_2\cdot \text{LReLU}(W_1 \cdot [u, \tilde{z}] + b_1) + b_2\Big) + b_3
	\end{split}
    \end{align}
where $W_1$, $W_2$, and $W_3$ are $2 \times 2$ weight matrices; $b_1$, $b_2$, and $b_3$ are $2 \times 1$ bias vectors.

The second class, which we denote as \textsc{AMLP} (Augmented MLP), has a multiplicative term as an augmented input unit. We expect that this multiplication term allows the vertical signal ($u^{(l+1)}$) to override the lateral signal ($\tilde{z}$), and also allows the lateral signal to select where the vertical signal is to be instantiated. Since the approximation of multiplication is not easy for a single-layer MLP, we explicitly add the multiplication term as an extra input to the combinator function. \textsc{AMLP} maps three scalars  $[u, \tilde{z}, u \odot \tilde{z}]$ to a single output. We use the same LReLU unit for \textsc{AMLP}.

We do similar experiments as in the \textsc{MLP} case and include results for $[4]$, $[2, 2]$ and $[2, 2, 2]$ hidden layer configurations.

Both \textsc{MLP} and \textsc{AMLP} weight parameters are randomly initialized to $\mathcal{N}(0, \eta)$. $\eta$ is considered to be a hyperparameter and tuned on the validation set. Precise values for the best $\eta$ values are listed in Appendix.
\subsection{Methodology}
The experimental setup includes two semi-supervised classification tasks with 100 and 1000 labeled examples and a fully-supervised classification task with 60000 labeled examples for Permutation-Invariant MNIST handwritten digit classification. Labeled examples are chosen randomly but the number of examples for different classes is balanced. The test set is not used during all the hyperparameter search and tuning. Each experiment is repeated 10 times with 10 different but fixed random seeds to measure the standard error of the results for different parameter initializations and different selections of labeled examples.

All variants and the vanilla Ladder Network itself are trained using the ADAM optimization algorithm \citep{ kingma2014adam} with a learning rate of 0.002 for 100 iterations followed by 50 iterations with a learning rate decaying linearly to 0. Hyperparameters including the standard deviation of the noise injection and the denoising weights at each layer are tuned separately for each variant and each experiment setting (100-, 1000-, and fully-labeled). Hyperparmeters are optimized by either a random search \citep{bergstra2012random}, or a grid search, depending on the number of hyperparameters involved (see Appendix for precise configurations).

\section{Results \& Discussion}
Table \ref{tab:results} collects all results for the variants and the baselines. The results are organized into five main categories for each of the three tasks. The \textsc{Baseline} model is a simple feed-forward neural network with no reconstruction penalty and \textsc{Baseline+noise} is the same network but with additive noise at each layer. The best results in terms of average error rate on the test set are achieved by the proposed \textsc{AMLP} combinator function: in the fully-supervised setting, the best average error rate is \textbf{$0.569\pm0.010$}, while in the semi-supervised settings with 100 and 1000 labeled examples, the averages are \textbf{$1.002\pm0.037$} and \textbf{$0.974\pm0.021$} respectively. A visualization of the a learned combinator function is shown in Appendix.

{
\centering
\begin{table*}[htb]
\centering
\caption{PI MNIST classification results for the vanilla Ladder Network and its variants trained on 100, 1000, and 60000 (full) labeled examples. AER and SE stand for Average Error Rate and its Standard Error of each variant over 10 different runs. \textsc{Baseline} is a multi-layer feed-forward neural network with no reconstruction penalty.}
  \begin{tabular}{ccccccc}
  \addlinespace
  \toprule
  {\bf } & \multicolumn{2}{c}{{\bf 100}} & \multicolumn{2}{c}{{\bf 1000}} & \multicolumn{2}{c}{{\bf 60000}}\\
  \midrule
  {\bf Variant} & {\bf AER (\%)} & {\bf SE} & {\bf AER (\%)} & {\bf SE} & {\bf AER (\%)} & {\bf SE}\\
Baseline & 25.804  & $\pm$ 0.40 & 8.734  & $\pm$ 0.058  & 1.182  & $\pm$ 0.010\\
Baseline+noise & 23.034  & $\pm$ 0.48 & 6.113  & $\pm$ 0.105  & 0.820  & $\pm$ 0.009\\
Vanilla & 1.086  & $\pm$ 0.023 & 1.017  & $\pm$ 0.017  & 0.608  & $\pm$ 0.013\\
\addlinespace
\midrule
\addlinespace
FirstNoise    & 1.856  & $\pm$ 0.193 & 1.381  & $\pm$ 0.029  & 0.732  & $\pm$ 0.015\\
FirstRecons   & 1.691  & $\pm$ 0.175 & 1.053  & $\pm$ 0.021  & 0.608  & $\pm$ 0.013\\
FirstN\&R   & 1.856  & $\pm$ 0.193 & 1.058  & $\pm$ 0.175  & 0.732  & $\pm$ 0.016\\
NoLateral   & 16.390  & $\pm$ 0.583 & 5.251  & $\pm$ 0.099  & 0.820  & $\pm$ 0.009\\
\addlinespace
\midrule
\addlinespace
RandInit   & 1.232  & $\pm$ 0.033 & 1.011  & $\pm$ 0.025  & 0.614  & $\pm$ 0.015\\
RevInit   & 1.305  & $\pm$ 0.129 & 1.031  & $\pm$ 0.017  & 0.631  & $\pm$ 0.018\\
NoSig   & 1.608  & $\pm$ 0.124 & 1.223  & $\pm$ 0.014  & 0.633  & $\pm$ 0.010\\
NoMult   & 3.041  & $\pm$ 0.914 & 1.735  & $\pm$ 0.030  & 0.674  & $\pm$ 0.018\\
Linear   & 5.027  & $\pm$ 0.923 & 2.769  & $\pm$ 0.024  & 0.849 & $\pm$ 0.014\\
\addlinespace
\midrule
\addlinespace
Gaussian   & 1.064  & $\pm$ 0.021 & 0.983  & $\pm$ 0.019  & 0.604  & $\pm$ 0.010\\
GatedGauss   & 1.308  & $\pm$ 0.038 & 1.094  & $\pm$ 0.016  & 0.632  & $\pm$ 0.011\\
\addlinespace
\midrule
\addlinespace
MLP\,$[4]$   & 1.374  & $\pm$ 0.186 & 0.996  & $\pm$ 0.028  & 0.605  & $\pm$ 0.012\\
MLP\,$[2,2]$   & 1.209  & $\pm$ 0.116 & 1.059  & $\pm$ 0.023  & 0.573  & $\pm$ 0.016\\
MLP\,$[2,2,2]$  & 1.274 & $\pm$ 0.067 & 1.095  & $\pm$ 0.053  & 0.602  & $\pm$ 0.010\\
AMLP\,$[4]$   & 1.072  & $\pm$ 0.015 & \textbf{0.974}  & $\pm$ 0.021  & 0.598  & $\pm$ 0.014\\
AMLP\,$[2,2]$   & 1.193  & $\pm$ 0.039 & 1.029  & $\pm$ 0.023  & \textbf{0.569}  & $\pm$ 0.010\\
AMLP\,$[2,2,2]$   & \textbf{1.002}  & $\pm$ 0.038 & 0.979  & $\pm$ 0.025  & 0.578  & $\pm$ 0.013\\
  \bottomrule
  
  \end{tabular}
\label{tab:results}
\end{table*}
}

\subsection{Variants derived by removal}
The results in the table indicate that in the fully-supervised setting, adding noise either to the first layer only or to all layers leads to a lower error rate with respect to the baselines. Our intuition is that the effect of additive noise to layers is very similar to the weight noise regularization method \citep{graves2011practical} and dropout~\citep{Hinton-et-al-arxiv2012-small}.

In addition, it seems that removing the lateral connections hurts much more than the absence of noise injection or reconstruction penalty in the intermediate layers. It is also worth mentioning that hyperparameter tuning yields zero weights for penalizing the reconstruction errors in all layers except the input layer in the fully-supervised task for the vanilla model. Something similar happens for \textsc{NoLateral} as well, where hyperparameter tuning yields zero reconstruction weights for all layers including the input layer. In other words, \textsc{NoLateral} and \textsc{Baseline+noise} become the same models for the fully-supervised task. Moreover, the weights for the reconstruction penalty of the hidden layers are relatively small in the semi-supervised task. This is in line with similar observations (relatively small weights for the unsupervised part of the objective) for the hybrid discriminant RBM~\citep{larochelle2008}.

The third part of Table \ref{tab:results} shows the relative performance of different combinator functions by removal. Unsurprisingly, the performance deteriorates considerably if we remove the sigmoid non-linearity (\textsc{NoSig}) or the multiplicative term (\textsc{NoMult}) or both (\textsc{Linear}) from the vanilla combinator. Judging from the size of the increase in average error rates, the multiplicative term is more important than the sigmoid unit.

\subsection{Variants derived by replacements}
As described in Section \ref{sec:vanilla_variants} and Equation \ref{vanilla_init}, the per-element weights of the lateral connections are initialized to ones while those of the vertical are initialized to zeros. Interestingly, the results are slightly worse for the \textsc{RandInit} variant, in which these weights are initialized randomly. The \textsc{RevInit} variant is even worse than the random initialization scheme. 
We suspect that the reason is that the optimization algorithm finds it easier to reconstruct a representation $z$ starting from its noisy version $\tilde{z}$, rather than starting from an initially arbitrary reconstruction from the untrained upper layers. Another justification is that the initialization scheme in Equation \ref{vanilla_init} corresponds to optimizing the Ladder Network as if it behaves like a stack of decoupled DAEs initially, therefore during early training it is like that the Auto-Encoders are trained more independently.

The \textsc{Gaussian} combinator performs better than the vanilla combinator. \textsc{GatedGauss}, the other variant with strict $0 < \sigma(u) < 1$, does not perform as well as the one with unconstrained $\sigma(u)$. In the \textsc{Gaussian} formulation, $\tilde{z}$ is regulated by two functions of $u$: $\mu(u)$ and $\sigma(u)$. This combinator interpolates between the noisy activations and the predicted reconstruction, and the scaling parameter can be interpreted as a measure of the certainty of the network.

Finally, the \textsc{AMLP} model yields state-of-the-art results in all of 100-, 1000- and 60000-labeled experiments for PI MNIST. It outperforms both the \textsc{MLP} and the vanilla model. The additional multiplicative input unit $\tilde{z} \odot u$ helps the learning process significantly.
% Furthermore, \textsc{AMLP} combinators are arguably more intuitive than the unusual vanilla  design (equation \ref{vanilla_comb}).

\subsection{Probabilistic Interpretations of the Ladder Network}

Since many of the motivations behind regularized autoencoder architectures are based on observations about generative models, we briefly discuss how the Ladder Network can be related to some other models with varying degrees of probabilistic interpretability. Considering that the components that are most defining of the Ladder Network seem to be the most important ones for semi-supervised learning in particular, comparisons with generative models are at least intuitively appealing to get more insight about how the model learns about unlabeled examples. 

By training
the individual denoising autoencoders that make up the Ladder Network with a single objective function, this coupling goes as far as encouraging the lower
levels to produce representations that are going to be easy to reconstruct by the upper levels.
We find a similar term (-log of the top-level prior evaluated at the output of the encoder)
in hierarchical extensions of the variational autoencoder~\citep{Rezende-et-al-ICML2014,Bengio-arxiv2014}. 
While the Ladder Network differs too much from an actual variational autoencoder to be treated as such, the similarities can still give one intuitions about the role of the noise and the interactions between the layers. Conversely, one also may wonder how a variational autoencoder might benefit from some of the components of Ladder Network like Batch Normalization and multiplicative connections.

When one simply views the Ladder Network as a peculiar type of denoising autoencoder, one could extend the recent work on the generative interpretation
of denoising autoencoders~\citep{Alain+Bengio-ICLR2013-small,Bengio-et-al-NIPS2013-small} to interpret the Ladder Network as a generative model as well.
It would be interesting to see if the Ladder Network architecture can be used to generate samples and if the architecture's success at semi-supervised learning translates to this profoundly different use of the model.

\section{Conclusion}
The paper systematically compares different variants of the recent Ladder Network architecture \citep{Rasmus2015,valpola2014neural} with two feedforward neural networks as the baselines and the standard architecture (proposed in the original paper).
Comparisons are done in a deconstructive way, starting from the standard architecture. Based on the comparisons of different variants we conclude that:
\begin{itemize}
\item Unsurprisingly, the reconstruction cost is crucial to obtain the desired regularization from unlabeled data.
\item Applying additive noise to each layer and especially the first layer has a regularization effect which helps generalization. This seems to be one of the most important contributors to the performance on the fully supervised task.
\item The lateral connection is a vital component in the Ladder architecture 
to the extent that removing it considerably deteriorates the performance for all of the semi-supervised tasks.
\item The precise choice of the combinator function has a less dramatic impact, although the vanilla combinator can be replaced by the Augmented MLP to yield better performance, in fact allowing us to improve the record error rates on Permutation-Invariant MNIST for semi- and fully-supervised settings.
\end{itemize}
We hope that these comparisons between different architectural choices will help to improve understanding of semi-supervised learning's success for the Ladder Network and like architectures, and perhaps even deep architectures in general.

\nocite{langley00}

\bibliography{example_paper}
\bibliographystyle{icml2016}

\end{document}